\documentclass[conference]{IEEEtran}

\usepackage{makecell}
\usepackage{cite}
\usepackage{graphicx}
\usepackage{amsmath,amssymb,amsfonts}
\usepackage{algorithmic}
\usepackage{textcomp}
\usepackage{xcolor}
\usepackage[ruled,vlined]{algorithm2e}
\usepackage{xurl}
\usepackage{multirow}
\usepackage{array}
\usepackage{booktabs}
\usepackage{caption}
\usepackage[most]{tcolorbox}
\usepackage{xcolor}
\usepackage{amsmath}
\usepackage{xcolor}
\definecolor{myblue}{rgb}{0, 0.44, 0.75} 

\usepackage[
    colorlinks=true,
    citecolor=myblue,
    linkcolor=black,
    urlcolor=myblue
]{hyperref}

\definecolor{mygray}{rgb}{0.95,0.95,0.95}
\definecolor{darkgray}{rgb}{0.2,0.2,0.2}
\definecolor{mypurple}{rgb}{0.4,0.2,0.6}
\definecolor{mygreen}{rgb}{0.0,0.5,0.3}

\title{Enhancing Spatial Reasoning in Vision-Language Models via Chain-of-Thought Prompting and Reinforcement Learning}

\author{
    \IEEEauthorblockN{Binbin Ji}
    \IEEEauthorblockA{
        \textit{Courant} \\
        \textit{New York University} \\
        New York, United States \\
        bj2414@nyu.edu
    }
    \and
    \IEEEauthorblockN{Siddharth Agrawal}
    \IEEEauthorblockA{
        \textit{Courant} \\
        \textit{New York University} \\
        New York, United States \\
        sa6740@nyu.edu
    }
    \and
    \IEEEauthorblockN{Qiance Tang}
    \IEEEauthorblockA{
        \textit{Courant} \\
        \textit{New York University} \\
        New York, United States \\
        qt2094@nyu.edu
    }
    \and
    \IEEEauthorblockN{Yvonne Wu}
    \IEEEauthorblockA{
        \textit{Courant} \\
        \textit{New York University} \\
        New York, United States \\
        yw4142@nyu.edu
    }
}

\begin{document}

\maketitle

\begin{abstract}
This study investigates the spatial reasoning capabilities of vision-language models (VLMs) through Chain-of-Thought (CoT) prompting and reinforcement learning. We begin by evaluating the impact of different prompting strategies and find that simple CoT formats—where the model generates a reasoning step before the answer—not only fail to help, but can even harm the model’s original performance. In contrast, structured multi-stage prompting based on scene graphs (SceneGraph CoT) significantly improves spatial reasoning accuracy. To further improve spatial reasoning ability, we fine-tune models using Group Relative Policy Optimization (GRPO) on the SAT dataset and evaluate their performance on CVBench. Compared to supervised fine-tuning (SFT), GRPO achieves higher accuracy on Pass\@1 evaluations and demonstrates superior robustness under out-of-distribution (OOD) conditions. In particular, we find that SFT overfits to surface-level linguistic patterns and may degrade performance when test-time phrasing changes (e.g., from “closer to” to “farther from”). GRPO, on the other hand, generalizes more reliably and maintains stable performance under such shifts. Our findings provide insights into how reinforcement learning and structured prompting improve the spatial reasoning capabilities and generalization behavior of modern VLMs. All code is open source at \href{https://github.com/Yvonne511/spatial-vlm-investigator/}{GitHub} to facilitate further research, transparency, and open science.

\end{abstract}

\section{Introduction}

Vision-language models (VLMs) have achieved impressive success across a wide range of multimodal tasks, including image captioning, visual question answering (VQA), and image-text retrieval. Despite this progress, spatial reasoning—understanding object locations, relative depth, geometric relations, and spatial alignment—remains a significant challenge for these models. Robust spatial reasoning is critical for real-world deployment of VLMs in domains such as robotics, autonomous navigation, and human-object interaction. A major difficulty lies in enabling VLMs to generalize beyond training distributions. This fragility highlights the need for strategies that can improve generalization, such as structured Chain-of-Thought prompting or more robust training methodologies.

Recent efforts to enhance reasoning capabilities in large language models have explored Chain-of-Thought (CoT) prompting, where models are encouraged to generate intermediate reasoning steps before producing final answers. While this strategy has shown success in math and logical reasoning, our empirical results show that naïve CoT prompting (e.g., "think first, then answer") not only fails to improve spatial reasoning in VLMs, but may even degrade performance. In contrast, structured multi-stage prompting that incorporates scene-level relational information—such as Scene Graph–based CoT—can significantly improve spatial understanding.

To further address vision-language misalignment, we adopt Group Relative Policy Optimization (GRPO)\cite{shao2024deepseekmathpushinglimitsmathematical}, a reinforcement learning algorithm that optimizes model behavior through relative comparisons of generated responses. We fine-tune models such as PaLI-Gemma2 using GRPO on the SAT dataset and evaluate generalization on CVBench. In comparison to supervised fine-tuning (SFT), GRPO yields higher Pass@1 performance and exhibits greater robustness under out-of-distribution (OOD) evaluations, including semantic polarity shifts.

We conduct detailed evaluations across multiple benchmarks (VSR\cite{liu2023visualspatialreasoning},SAT\cite{ray2025satdynamicspatialaptitude}, CVBench\cite{tong2024cambrian1fullyopenvisioncentric}, CLEVR-CoGenT\cite{johnson2017clevr}) and analyze generalization behavior under controlled linguistic variation and across different prompting strategies.

\section{Background and Related Work}


\subsection{RL Finetuned VLM}
Reinforcement learning (RL) has been explored as a powerful tool for improving the generalization capabilities of visual language models (VLMs). 
Visual-RFT proposed a method to leverage RL for fine-tuning VLMs to perform segmentation tasks, showing promising results in visual representation learning \cite{liu2025visualrftvisualreinforcementfinetuning}. RL also provides better generalization ability. Chu et al. and Chen et al. demonstrated that RL fine-tuning can significantly enhance generalization in VLMs compared to supervised fine-tuning (SFT)\cite{chu2025sftmemorizesrlgeneralizes}. R1-V then used GRPO to fine-tune the small Qwenvl model on the CLEVR data set, proving the reasoning and generalization ability of GRPO in the spatial reasoning task\cite{chen2025r1v,chen2025rlvrinvlms}.Building on these findings, many works like VLM-R1 extended GRPO’s application to broader spatial reasoning benchmarks and larger vision-language models, showing that GRPO consistently improves model alignment, robustness, and task generalization across various spatial tasks\cite{zhou2025r1zerosahamomentvisual,shen2025vlmr1stablegeneralizabler1style,li2025thinkthinkstudyexplicit}. After its success in image-based spatial reasoning, GRPO has been increasingly applied to video-based spatial tasks to handle temporal dynamics and object interactions over time following the spatial task dimensions proposed in “Thinking in Space”\cite{feng2025videor1reinforcingvideoreasoning,ouyang2025spatialr1enhancingmllmsvideo,liao2025improvedvisualspatialreasoningr1zerolike,yang2024thinkingspacemultimodallarge}.

\subsection{Spatial CoT}
Spatial reasoning in visual language models has been the focus of multiple studies. CoT has been considered as a method that utilizes reasoning from VLM for better spatial understanding.
Scene Graph-based Chain-of-Thought (Scene Graph CoT) involves structured reasoning using scene graphs generated from visual inputs to facilitate spatial reasoning.

Compositional CoT is structured into two steps: scene graph generation and answer generation, reducing the likelihood of reward hacking \cite{mitra2024compositionalchainofthoughtpromptinglarge}. Optical Flow CoT extends Scene Graph CoT to handle dynamic spatial reasoning tasks, particularly in benchmarks such as SAT \cite{liu2025spatialcotadvancingspatialreasoning}. This approach encodes positional differences to track egocentric and object-centric movement. Few-shot prompting aims to mitigate model shortcutting by separating scene graph generation and answer prediction into distinct tuples \cite{wu2024minds}. Additionally, conversational prompting reformulates few-shot examples as user-assistant interactions to promote structured reasoning.

There are additional methods that could be applied to an existing CoT format. Program-of-Thought (PoT) uses logical operators on top of scene graphs to guide reasoning \cite{liu2025spatialcotadvancingspatialreasoning}. Symbols-of-Thought (SoT) encodes visual elements as symbolic tokens \cite{hu2023chain}. Visualization-of-Thought (VoT) utilizes visual guides to reinforce reasoning \cite{wu2024minds}.

There are also other methods that train the model to include a knowledge framework in the CoT promoting answer. SpatialCoT introduced a method for coordinating language alignment and CoT to improve task planning \cite{liu2025spatialcotadvancingspatialreasoning}. Paper, Thinking in Space, further explored the concept by prompting multimodal large language models (MLLMs) to express their internal spatial representations using cognitive maps \cite{yang2024thinkingspacemultimodallarge}.

\subsection{VLM Spatial Reasoning Benchmark}
There has been extensive research on the spatial reasoning capabilities of vision-language models, with various studies proposing different evaluation criteria and benchmarks.

CLEVR introduced a simulated and diagnostic dataset containing diverse 3D shapes, object colors, and textures \cite{mminstructionclevrcogentvalb}. The dataset includes questions that focus on attribute identification, counting, comparison, multi-object attention, and logical operations. Its primary objective is to provide a benchmark free from strong biases that models could exploit to artificially boost performance. Building on CLEVR, the Super-CLEVR dataset extends the original by incorporating out-of-distribution data, more complex shapes, to assess model generalization \cite{li2023superclevrvirtualbenchmarkdiagnose}.

In contrast, the Visual Spatial Reasoning (VSR) dataset seeks to enhance the expressiveness of spatial relations \cite{liu2023visualspatialreasoning}. Unlike CLEVR and its derivatives, which include only four relations — “left,” “right,” “behind,” and “in front of” — VSR incorporates 66 spatial relations from real-world images, expanding the complexity and scope of spatial reasoning evaluation. However, since VSR employs a true/false evaluation format, the implications of achieving accuracy above 50\% remain ambiguous, as it is unclear whether such accuracy reflects genuine spatial understanding or mere chance.

While these datasets focus on spatial reasoning in controlled or simulated settings, CV-Bench adopts a more focused approach by zooming into the vision capabilities of VLMs and isolating the effects of the language model backbone. CV-Bench, introduced by Cambrian-1, categorizes tasks into four distinct groups: Spatial Relationship, Object Counting, Depth Order, and Relative Distance \cite{tong2024cambrian1fullyopenvisioncentric}.


\section{Methodology}


This paper explores VLMs' spatial reasoning through various metrics. The methodology is structured into three primary phases: Dataset Selection and Benchmarking, GRPO Implementation and Fine-Tuning, and Scene Graph-Based CoT Prompting.

\subsection{Dataset Selection and Benchmarking}
We employ a diverse set of datasets to evaluate model performance on spatial reasoning tasks, including Clevr\_CoGenT\_ValB \cite{mminstructionclevrcogentvalb}, Super-CLEVR \cite{li2023superclevrvirtualbenchmarkdiagnose}, the Visual Spatial Reasoning (VSR) dataset \cite{liu2023visualspatialreasoning}, and CV-bench \cite{tong2024cambrian1fullyopenvisioncentric}. These benchmarks are strategically selected to assess the model under distinct scenarios, such as GRPO fine-tuning and scene graph-based chain-of-thought (CoT) prompting.

Prior to implementing modifications, we conducted baseline evaluations to assess the model's inherent capabilities. Specifically, we utilized CLEVR CoGenT ValB, leveraging object-level information such as relations, directions, material, color, size, and shape to construct comprehensive prompts for each image-question pair \cite{johnson2017clevr}. 

The baseline results reveal a modest increase in accuracy when incorporating structured prompts:
\begin{itemize}
    \item Without prompt: 47.05\%
    \item With prompt: 48.82\%
    \item Prompt-only (removing image): 52.25\%
\end{itemize}

The marginal improvement when using the prompt alongside the image suggests that the model is moderately responsive to explicit contextual information but may struggle with integrating visual and textual data effectively. The higher accuracy with prompt-only conditions indicates a potential over-reliance on linguistic cues, underscoring a limitation in the model's visual reasoning capabilities. Additionally, the relatively low accuracy even with comprehensive image coverage in the prompts further underscores the model’s constrained reasoning capacity.

To further probe the robustness of the model, we experimented with a variety of prompt formulations. However, these alterations yielded negligible improvements in accuracy, reinforcing the notion that the model may be constrained by its inherent architecture.

These findings indicate that while prompting strategies can provide incremental gains, substantial performance enhancements may require more targeted training approaches. Further analysis will investigate the impact of scene graph-based CoT prompting and GRPO fine-tuning to systematically assess these potential avenues.

\subsection{GRPO Implementation and Fine-Tuning}
We implement GRPO, fine-tuning on the SAT dataset to assess its efficacy in enhancing counting accuracy and spatial reasoning \cite{ray2025satdynamicspatialaptitude}. The model is then tested on CVBench to evaluate its generalization across datasets and task variations.

The GRPO algorithm optimizes the VLM by integrating reinforcement learning with reward computation and gradient scaling. The GRPO loss is defined as follows:
\begin{align}
L_{GRPO} = - \mathbb{E}_{\pi} & \left[ \exp(\log \pi(a|s) - \log \pi_{ref}(a|s)) \right. \notag \\
& \left. \cdot A(s, a) - \beta \cdot KL(\pi(a|s) \,||\, \pi_{ref}(a|s)) \right]
\end{align}

$\pi(a|s)$ is the model's probability distribution over actions $a$ given state $s$; $\pi_{ref}(a|s)$ is the reference model's probability distribution; $A(s, a)$ is the advantage function; $\beta$ is a scaling factor for the KL divergence.

We investigate prompting strategies to improve Visual Language Models (VLMs) on a diverse suite of spatial reasoning tasks drawn from CVBench, SAT, VSR, and Clevr-CoGenT benchmarks. Our methodology primarily explores \textbf{Scene Graph-based Chain-of-Thought (Scene Graph CoT)} reasoning, with ablations comparing its efficacy against other reasoning paradigms such as Optical Flow CoT, Program-of-Thought (PoT), Symbols-of-Thought (SoT), and Visualization-of-Thought (VoT).

\subsection{Scene Graph CoT Prompting Framework}

We formalize a two-step prompting strategy involving distinct model queries:

\subsubsection*{Step 1: Scene Graph Generation}

Let \( I \) be the input image and \( Q \) be the associated question. The first prompt \( P_{\text{SG}} \) requests the model to produce a scene graph \( R_{\text{SG}} \in \mathcal{T}_{\text{SG}} \), structured as JSON and containing:

\begin{itemize}
    \item \( O = \{o_1, o_2, \ldots, o_n\} \): relevant objects.
    \item \( A = \{a_1, a_2, \ldots, a_n\} \): object attributes.
    \item \( R = \{r_{ij}\} \): relationships among objects.
\end{itemize}

The user prompt \( P_{\text{SG}} \) is:
\begin{quote}
\textit{``For the provided image and its associated question, think and generate a scene graph in JSON format that includes the following: (1) Objects relevant to answering the question, (2) Object attributes relevant to answering the question, (3) Object relationships relevant to answering the question.''}
\end{quote}

The corresponding system prompt is:
\begin{quote}
\textit{``You are an AI assistant proficient in visual and spatial reasoning with tasks involving counting, relations, depth, distances, etc., and generate scene graphs based on images and questions. Think and then answer.''}
\end{quote}

We extract the scene graph \( SG \subseteq R_{\text{SG}} \) using regex rules, prioritizing code blocks (\texttt{\char`\`}\texttt{\char`\`}\texttt{\char`\`}), JSON braces (\{ \}), or fallback tags (\texttt{<code>}...\texttt{</code>}). If all fail, we set \( SG = R_{\text{SG}} \).

\subsubsection*{Step 2: Answer Generation Using Scene Graph}

In the second step, the model receives \( I, Q, SG \) and is prompted with \( P_R \):

\begin{quote}
\textit{``Use the image and scene graph as context and answer the following question.''}
\end{quote}

The system prompt is:
\begin{quote}
\textit{``You are an AI assistant proficient in visual and spatial reasoning with tasks involving counting, relations, depth, distances, etc. Think and then answer. Final answer should be provided between \texttt{<answer>} and \texttt{</answer>} tags. \{Further requirements for automated evaluation go here\}''}
\end{quote}

Although step 1 is only required to generate the scene graph, the model frequently also includes the correct answer implicitly, while step 2 mainly serves to structure the output for evaluation.

\subsection{Optical Flow CoT}

For dynamic spatial reasoning tasks in the SAT benchmark, especially \textbf{Egocentric Movement} and \textbf{Object Movement}, we employ \textbf{Optical Flow CoT}. Given two successive images \( I_1, I_2 \), the model is tasked to identify motion (left/right) either in the egocentric frame (camera) or of specific objects.

This is restructured as a special case of Scene Graph CoT, where the scene graph encodes positional differences:
\[
\Delta x(o_i) = 
\begin{cases}
``left" & \text{if object } o_i \text{ moved left} \\
``stationary" & \text{if object } o_i \text{ remained stationary} \\
``right" & \text{if object } o_i \text{ moved right}
\end{cases}
\]

\subsection{Few-Shot and Conversational Prompting}

To further enhance performance, we explore in-context learning via \textbf{k-shot prompting}. Our prompting strategy builds on the observation that jointly asking for both a scene graph and an answer in a single prompt significantly degrades accuracy. We attribute this to \textit{reward hacking}, where the model learns to shortcut reasoning by jumping to the answer, often at the cost of correctness. Consequently, we adopt a two-stage prompting scheme, which we extend into the few-shot setting.

Each example in the few-shot context is decomposed into two tuples:

\begin{align*}
T_{i,1} &= (I_i, Q_i, R_i, SG_i) \quad \text{(Scene Graph generation)} \\
T_{i,2} &= (I_i, Q_i, R_i, SG_i, A_i) \quad \text{(Answer generation)}
\end{align*}

Where:
\begin{itemize}
    \item \( I_i \): Input image
    \item \( Q_i \): Natural language question
    \item \( R_i \): Intermediate reasoning (optional or implicit)
    \item \( SG_i \): Scene graph representing relevant objects, attributes, and relationships
    \item \( A_i \): Final answer
\end{itemize}

\subsubsection*{Structured Few-Shot Prompting}

In this method, we emulate the two-stage pipeline explicitly. For the scene graph generation phase (Step 1), we provide \( k \) examples of \( T_{i,1} \), deliberately omitting the answer \( A_i \) to avoid premature prediction. This encourages the model to generate structured representations without jumping to conclusions. For answer generation (Step 2), we switch to providing full examples \( T_{i,2} \) where the answer is included, guiding the model toward completing the chain of reasoning.

This decoupling proved necessary. Early experiments with flat tuple formats (i.e., providing full examples of \( T_i = (I_i, Q_i, R_i, SG_i, A_i) \)) led to the model copying the few-shot examples almost verbatim—merely swapping in a different answer while reusing previous scene graphs. This behavior mirrored the issues observed in single-step prompting, deteriorating overall accuracy.

\subsubsection*{Conversational Few-Shot Prompting}

To further mitigate rigid behavior and induce more flexible reasoning, we reformulate few-shot prompting in a chat-like format, which we refer to as \textbf{k-shot Conversational Prompting}.

In this approach, few-shot examples are represented as alternating user-assistant exchanges $\forall i=0 \cdots k$:
\begin{itemize}[noitemsep]
    \item \textbf{User}: provides \( (I_i, Q_i) \)
    \item \textbf{Assistant}: pre-intialised with \( (R_i, SG_i) \)
\end{itemize}

It is important to note that the Assistant's responses \( (R_i, SG_i) \) are not generated by the VLM but provided by the user for k-shot prompting and in-context learning. Still, the messages are classified as ones generated from the assistant within the chat history. 

The actual prompt concludes with the new question appended to this dialogue history:
\begin{quote}
\texttt{User: Image: \dots{} \\ Question: \dots{} \\ Assistant: Scene Graph:}
\end{quote}

This naturalistic interaction style biases the model to imitate thoughtful, structured behavior and generalize beyond the specifics of the few-shot examples. After the model produces a scene graph, the answer phase is initiated with additional user-assistant pairs that include full \( T_{i,2} \) tuples, ending with:
\begin{quote}
\texttt{User: Image: \dots{} \\ Question: \dots{} \\ Assistant: Answer:}
\end{quote}

We found this conversational method not only improved flexibility but also better aligned with the capabilities of instruction-tuned VLMs, which are pretrained to operate in chat-based environments. By explicitly separating scene understanding from answer prediction, we observed more faithful reasoning and reduced shortcutting.

\subsection{Describe CoT}
The CoT prompt here followed a 2 prompt strategy where the model
was prompted to: ``\textit{Describe the objects in this image: their color, shape,
size, location, relative location, reflectance and material (glossy is metallic,
matte is rubber).}'' in the first phase, and then in the second phase, to answer the question given the image and this information from the first prompt.

\subsection{Other CoT methods}
We conducted additional evaluations of alternative chain-of-thought (CoT) reasoning approaches, including Program-of-Thought \cite{liu2025spatialcotadvancingspatialreasoning}, Chain-of-Symbols \cite{hu2023chain}, and Visualization-of-Thought \cite{wu2024minds}. However, each of these methods consistently underperformed, yielding results that were not only inferior to Scene Graph CoT but also frequently worse than direct-answering baselines. These approaches often introduced unnecessary complexity or misaligned reasoning steps that degraded overall performance. Furthermore, we experimented with adding depth map information to Scene Graph prompts by utilising depth maps generating using DepthAnythingV2 \cite{depth_anything_v2}. 

\subsubsection*{Implementation Details}

For CVBench, we employ 4-shot prompting, including one representative task each from Counting, Relation, Depth, and Distance. For SAT, we add a fifth example for Optical Flow CoT, covering dynamic spatial reasoning tasks such as Egocentric Movement and Object Movement. All few-shot exemplars were curated using GPT-4o and chosen from high-confidence ground-truth alignments. The original Scene Graph CoT method was employed on Qwen2.5-VL-7B-Instruct and Llama-4-Scout-17B-16E-Instruct. The k-shot Scene Graph CoT method was employed on Llama-4-Scout-17B-16E-Instruct only.

\section{Results}
\subsection{Counting task and Finetuning with spatial data } 

The counting ability of a vision–language model (VLM) reflects not only its short-term memory and attention alignment capabilities \cite{johnson2017clevr}, but also its ability to spatially differentiate similar or closely positioned objects. While many counting tasks require the model to correctly focus on the relevant regions, accurate counting further depends on the model's capacity to segment and distinguish overlapping or adjacent instances of the same object category.

To investigate this, we conduct experiments on several counting-related benchmarks, including CLEVR\_CoGenT\_ValB, CV-Bench, Pixmo-Count \cite{deitke2024molmopixmoopenweights}, and Static Spatial Reasoning (SAT) \cite{ray2025satdynamicspatialaptitude}.

First, we prompt the model to solve a series of object-counting questions and visualize the corresponding token-level attention maps. These heatmaps show that, when attempting to count, the VLM often attends precisely to the correct object regions. However, despite accurate spatial focus, the model still occasionally produces incorrect counts. We hypothesize that these errors stem not from attention misplacement but from the encoder’s limited segmentation ability: it can locate objects but struggles to separate tightly clustered or overlapping instances.

This hypothesis is supported by the recent CLIP–SAM fusion work \cite{li2024clipsamclipsamcollaboration}, where the authors combine a CLIP-based VLM with a SAM segmentation backbone. Their results demonstrate that integrating CLIP's global feature alignment will cause imprecise segmentation. While combining CLIP and SAM makes the model performance improve on the location and boundary distinguish ability. In light of this, we further conjecture that enhancing our VLM encoder with segmentation-aware training should reduce “off-by-one” counting errors.

To test this, we evaluate performance using two complementary metrics on our counting dataset:
\begin{itemize}
    \item \textbf{Accuracy:} The percentage of predictions that exactly match the ground truth count.
    \item \textbf{Close-Call Percentage:} Defined as
    \begin{equation}
        \text{Close-Call \%} = \frac{\bigl|\{\,i : |\hat{y}_i - y_i| = 1 \text{ and } \hat{y}_i \neq y_i\}\bigr|}{\bigl|\{\,i : \hat{y}_i \neq y_i\}\bigr|} \times 100\%,
    \end{equation}
    where $\hat{y}_i$ is the model’s predicted count and $y_i$ is the ground truth. This metric isolates cases where the model’s prediction is only one off, revealing fine-grained segmentation failures.
\end{itemize}

By analyzing both metrics, we can distinguish two failure modes: (a) complete mislocalization or object-type confusion (low accuracy, low close-call rate) versus (b) fine-grained segmentation errors (moderate accuracy, high close-call rate). A high close-call percentage would support our hypothesis that, although the VLM correctly attends, it struggles to separate clustered instances—leading to “off-by-one” counting mistakes.

We test two variants of Qwen2.5-VL: the original Qwen2.5-VL-Instruct-3B (no spatial fine-tuning), and \textbf{SpaceQwenVL-2.5-3B}, which is fine-tuned on a SpatialVLM-generated dataset incorporating relative distances, depth cues, and explicit segmentation annotations. As shown in Table~\ref{tab:vertical-counting-results}, the segmentation-aware \textbf{SpaceQwenVL-2.5-3B} improves the accuracy of the model in most benchmarks, confirming the benefit of segmentation integration. Especially in CV-bench, a dataset that is proved to be more reliant on the visual information, the spatially-finetuned model is able to improve the accuracy by a 43.16\% margin.

\begin{table}[h]
    \centering
    \caption{Accuracy and close-call percentage (in parentheses) on counting tasks in these benchmarks.}
    \label{tab:vertical-counting-results}
    \renewcommand{\arraystretch}{1.3}
    \setlength{\tabcolsep}{1pt} 
    
    \begin{tabular}{l|c|c}
        \toprule
        \textbf{Dataset ($\le 1$ ratio)} & \textbf{Regular QwenVL2.5-3B} & \textbf{SpaceQwenVL2.5-3B} \\
        \midrule
        pixmo-count & 44.81\% (63.42\%) & \textbf{45.37\%} (60.34\%) \\

        Clevr\_CoGenT\_ValB & \textbf{81.10\%} (96.81\%) & 79.46\% (99.47\%) \\

        CVBench & 21.05\% (56.67\%) & \textbf{64.21\%} (72.34\%) \\

        SAT & 70.49\% (91.11\%) & \textbf{76.6\%} (100\%) \\
        \bottomrule
    \end{tabular}
\end{table}

\subsection{GRPO Fine-tuning Result}
Both prior work\cite{yang2024thinkingspacemultimodallarge} and our preliminary experiments suggest that sampled Chain-of-Thought (CoT) prompting does not effectively activate the spatial reasoning capabilities of vision-language models (VLMs). Therefore, in this work, we adopt a \textbf{"No-Thinking" strategy}, where the model is trained to directly generate the final answer without intermediate CoT reasoning.
We fine-tuned the PaLI-gemma2-3B-mix-224 model~\cite{steiner2024paligemma2familyversatile} using two GRPO configurations, summarized in Table~\ref{tab:grpo_configs}.  

\begin{table}[h]
    \vspace{-10pt}
    \centering
    \caption{GRPO Configurations}
    \label{tab:grpo_configs}
    \renewcommand{\arraystretch}{1.3}
    \setlength{\tabcolsep}{0.8pt} 
    
    \begin{tabular}{>{\arraybackslash}m{3cm}|>{\centering\arraybackslash}m{2.5cm}|>{\centering\arraybackslash}m{3cm}}
        \toprule
        \textbf{Configuration} & \textbf{GRPO-v1} & \textbf{GRPO-v2} \\
        \midrule
        LoRA Target Modules & \texttt{q\_proj, v\_proj} & \texttt{q\_proj, k\_proj, v\_proj, o\_proj, gate\_proj, up\_proj, down\_proj} \\
        Rank & 8 & 16 \\
        $\beta$ & 0.04 & 0.01 \\
        Temperature & 1.0 & 1.0 \\
        Additional Strategy & None & Dynamic Sampling \\
        \bottomrule
    \end{tabular}
    \begin{flushleft}
    \small{All models are based on the Pali-gemma2-3B-mix-224 \cite{steiner2024paligemma2familyversatile}}
    \end{flushleft}
    \vspace{-10pt}
\end{table}

\noindent\textbf{GRPO-v1: Initial Cross-modal Alignment.} 
GRPO-v1 is motivated by our empirical observation that a "describe-then-answer" generation pattern consistently improves performance on spatial reasoning tasks. This suggests that the vision encoder is capable of extracting meaningful visual information and that the language model can reason over such content effectively. However, the persistent errors observed in direct answer generation indicate a misalignment between visual and textual representations. To address this, GRPO-v1 applies LoRA-based fine-tuning specifically to the \texttt{q\_proj} and \texttt{v\_proj} layers—two critical components responsible for cross-modal projection—while keeping the rest of the model frozen. The result is shown in Table~\ref{tab:model_comparison}.

\noindent\textbf{GRPO-v2: Enhanced Integration and Dynamic Sampling.} 
While GRPO-V1 led to modest improvements (approximately 2\%), the gains were limited, suggesting that targeting only the \texttt{q\_proj} and \texttt{v\_proj} layers may not be sufficient to resolve the alignment issues between visual and language modalities. To enable deeper cross-modal integration, GRPO-V2 expands the set of LoRA-adapted modules, increases the LoRA rank, and decreases the $\beta$ value for GRPO.

\begin{figure}[!t]
    \centering
    \includegraphics[width=\columnwidth]{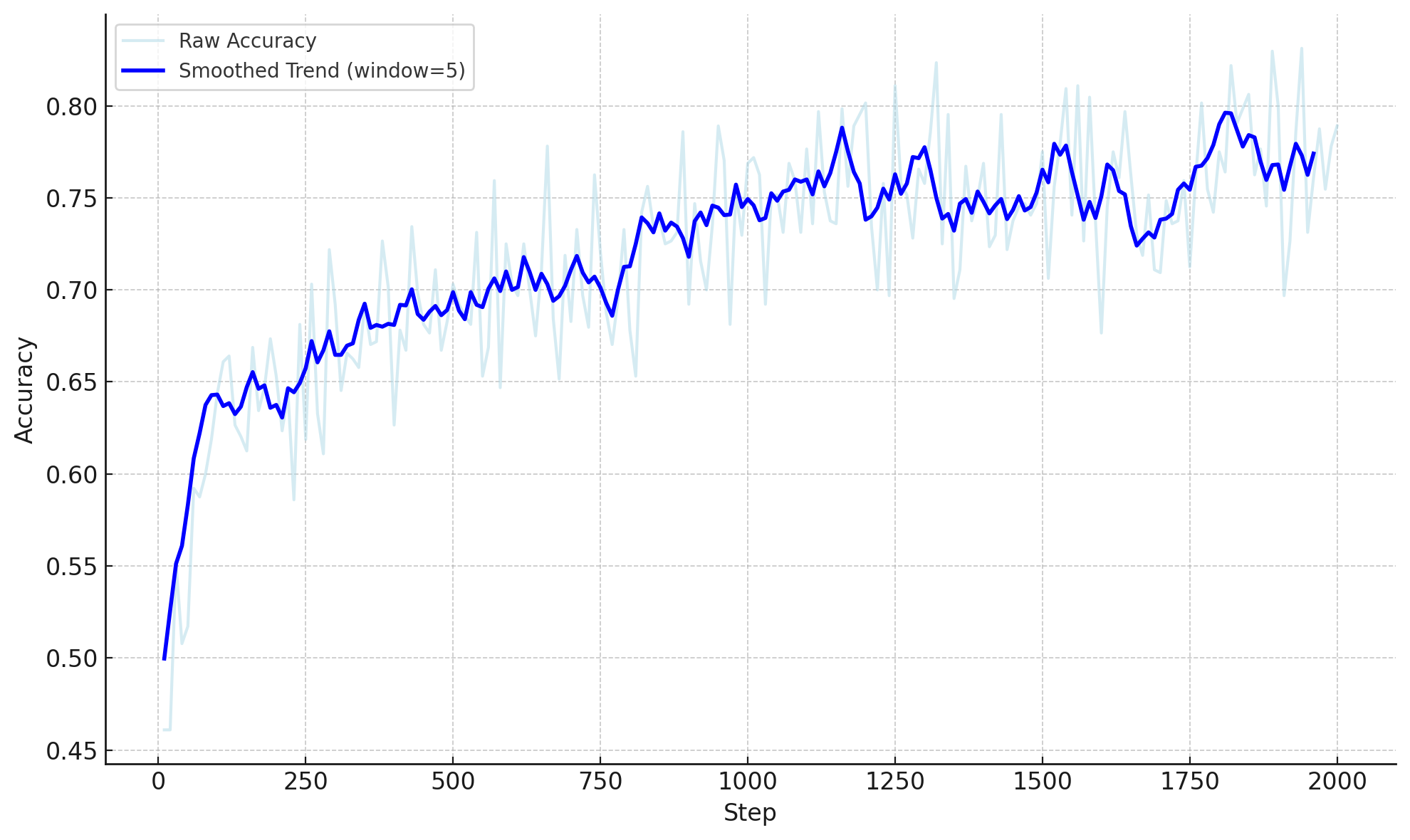}
    \caption{Accuracy Reward of GRPO-v2 model}
    \label{fig:grpo_acc}
\end{figure}

Moreover, during GRPO-V1 training, we observed a frequent collapse in group-wise reward signals: all generations within a group often received identical rewards. In such cases, the relative advantage signal—which is crucial for effective policy optimization—becomes zero, and the update direction of the model is driven almost entirely by the KL divergence term. This leads to unstable or regressive behavior, where the model is nudged back toward its pre-trained policy, potentially undoing the beneficial spatial alignment capability learned during fine-tuning. To address this issue, GRPO-V2 introduces \textbf{dynamic sampling strategy}, inspired by DAPO\cite{yu2025dapoopensourcellmreinforcement}, which focuses the learning process on truly informative examples - those where different responses actually receive different rewards. We found this approach especially important because reward collapse typically leads to decreased generation diversity and overly predictable outputs, further undermining the model's reasoning capabilities. The accuracy reward of this model is shown in Fig~\ref{fig:grpo_acc}.

\begin{table*}[!t]
    \centering
    \caption{Performance Comparison of Different Models on cvbench}
    \label{tab:model_comparison}
    \renewcommand{\arraystretch}{1.3}
    \begin{tabular}{l|c|c|c|c|c}
        \toprule
        \textbf{Model/Task} & \textbf{Counting} & \textbf{Relation} & \textbf{Depth} & \textbf{Distance} & \textbf{Total} \\
        & \textbf{(Pass@4)} & \textbf{(Pass@4)} & \textbf{(Pass@4)} & \textbf{(Pass@4)} & \textbf{(Pass@4)} \\
        \midrule
        Pali-gemma2-3B-mix-224 & 64\% & 77.08\% & 51.83\% & 14.83\% & 52.16\% \\
        Pali-gemma2-3B-mix-224-GRPO-v1 & 65.10\% & 78.41\% & 56.83\% & 18.00\% & 54.77\% \\
        Pali-gemma2-3B-mix-224-SFT & 65.07\%(72.08\%) & 80.92\%(82.92\%) & 61.67\%\textbf{(91.17\%)} & 59.50\%\textbf{(90.00\%)} & 66.79\%\textbf{(84.22\%)} \\
        Pali-gemma2-3B-mix-224-GRPO-v2 & \textbf{65.6\%(73.07\%)} & \textbf{84.92\%(86.00\%)} & \textbf{76.33\%}(88.17\%) & \textbf{62.67\%}(78.00\%) & \textbf{72.38\%}(81.31\%)\\
        \begin{tabular}[c]{@{}l@{}}Pali-gemma2-3B-mix-224 \\ + SceneGraph CoT\end{tabular} & 36.79\% & 62.13\% & 63.00\% & 52.50\% & 53.57\% \\
         \begin{tabular}[c]{@{}l@{}}Pali-gemma2-3B-mix-224-SFT \\ + SceneGraph CoT\end{tabular} & 40.13\% & 66.00\% & 53.50\% & \textbf{68.67\%} & 57.07\% \\
        \midrule
        Qwen2.5-VL-7B-Instruct & 50.76\% & 63.23\% & 66.33\% & 64.67\% & 60.54\% \\
        \begin{tabular}[c]{@{}l@{}}Qwen2.5-VL-7B-Instruct \\ + SceneGraph CoT\end{tabular} & \textbf{64.21\%} & \textbf{73.69\%} & \textbf{66.17\%} & \textbf{69.00\%} & \textbf{68.08\%} \\
        \midrule
        Llama-4-Scout-17B-16E-Instruct & 54.35\% & 55.08\% & 74.00\% & 31.33\% & 53.76\% \\
        \begin{tabular}[c]{@{}l@{}}Llama-4-Scout-17B-16E-Instruct \\ + SceneGraph CoT\end{tabular} & 67.64\% & 74.92\% & 89.17\% & 73.50\% & 75.66\% \\
        \begin{tabular}[c]{@{}l@{}}Llama-4-Scout-17B-16E-Instruct \\ + 4-shot CoT Prompting (SG + A split)\end{tabular} & 68.12\% & 75.88\% & 90.25\% & 74.42\% & 77.31\% \\
        \begin{tabular}[c]{@{}l@{}}Llama-4-Scout-17B-16E-Instruct \\ + 4-shot Conversational CoT Prompting\end{tabular} & \textbf{68.45\%} & \textbf{76.42\%} & \textbf{91.02\%} & \textbf{75.17\%} & \textbf{77.69\%} \\
        \midrule
        \begin{tabular}[c]{@{}l@{}}Llama-4-Scout-17B-16E-Instruct \\ + SceneGraph CoT + Depth Maps\end{tabular} & 46.97\% & 59.11\% & \textbf{89.38\%} & 71.29\% & 70.95\% \\
        \begin{tabular}[c]{@{}l@{}}Llama-4-Scout-17B-16E-Instruct \\ + SceneGraph CoT + PoT\end{tabular} & 49.12\% & 54.37\% & 65.84\% & 36.91\% & 51.56\% \\
        \begin{tabular}[c]{@{}l@{}}Llama-4-Scout-17B-16E-Instruct \\ + SceneGraph CoT + CoS\end{tabular} & 46.58\% & 51.43\% & 60.12\% & 33.87\% & 48.25\% \\
        \begin{tabular}[c]{@{}l@{}}Llama-4-Scout-17B-16E-Instruct \\ + SceneGraph CoT + VoT\end{tabular} & 47.39\% & 50.11\% & 58.63\% & 35.24\% & 47.84\% \\
        \begin{tabular}[c]{@{}l@{}}Llama-4-Scout-17B-16E-Instruct \\ + SceneGraph CoT + Confidence Scores\end{tabular} & 66.31\% & 74.01\% & 88.02\% & 71.84\% & 74.38\% \\
        \begin{tabular}[c]{@{}l@{}}Llama-4-Scout-17B-16E-Instruct \\ + SceneGraph CoT + VLBA Alignment\end{tabular} & 67.12\% & 73.44\% & 87.55\% & 72.20\% & 74.57\% \\
        \bottomrule
    \end{tabular}
    \begin{flushleft}
    \footnotesize
    \textbf{Note:} All results are reported as Pass@1 accuracy, unless specified in parentheses, which are Pass@4 values. \\
    PoT: Program-of-Thought prompting \cite{chen2022program}, CoS: Chain-of-Symbol prompting \cite{hu2023chain}, \\ VoT: Visualization-of-Thought prompting \cite{wu2024minds}. VLBA: Vision-Language Behavior Alignment \cite{liu2025spatialcotadvancingspatialreasoning}.\\ 
    SG + A Split: k-shot prompting where the first phase just has scene graphs, and the second phase has both Scene Graphs and Answers. As describe in Methodology section.
    \end{flushleft}. 
\end{table*}

\noindent\textbf{Comparison with Supervised Fine-tuning.} 
To further contextualize the impact of GRPO-based fine-tuning, we trained an additional baseline model using standard supervised fine-tuning (SFT) on the same spatial reasoning dataset. As shown in Table~\ref{tab:model_comparison}, both GRPO and SFT are effective in improving the spatial reasoning ability of the base PaLI-gemma2 model. However, we observe a \textbf{trade-off between different evaluation protocols}: SFT outperforms GRPO under the Pass@4 metric, whereas GRPO yields higher accuracy under Pass@1, which aligns with findings from prior work\cite{shao2024deepseekmathpushinglimitsmathematical}, and supports the view that GRPO primarily improves reasoning alignment by pruning suboptimal response paths, rather than by acquiring entirely new capabilities \cite{yue2025doesreinforcementlearningreally}. 

\noindent\textbf{Out-of-Distribution Generalization.} 
To further probe the generalization ability of each model, we designed an out-of-distribution (OOD) evaluation by altering the linguistic polarity of the test queries. While the training and original test sets primarily use "close to" phrasing (e.g., Which object is closer to the camera?), we constructed a variant where questions are rephrased using "far from" language (e.g., Which object is farther from the camera?), keeping the visual input unchanged. This setup allows us to assess whether the models have learned semantically spatial reasoning or merely memorized surface patterns.

\begin{table*}[!t]
    \centering
    \caption{Performance Comparison of Models on In-distribution and Out-of-distribution Tasks}
    \label{tab:ood_comparison}
    \renewcommand{\arraystretch}{1.3}
    
    \begin{tabular}{l|c|c|c|c|c}
        \toprule
        \textbf{Model} & \textbf{Depth} & \textbf{Depth\_OOD} & \textbf{Distance} & \textbf{Distance\_OOD} & \textbf{ID-OOD Gap (Distance)} \\
        \midrule
        Base Model & 51.83\% & 59.00\% & 14.83\% & 5.83\% & 9.00\% \\

        SFT Model & 61.67\% & {\color{red}56.33\% (-4.5\%)} & 59.50\% & 47.47\% & {\color{red}12.03\%} \\

        GRPO Model & 76.33\% & {\color{green}70.50\% (+19.5\%)} & 62.67\% & 59.50\% & {\color{green}3.17\%} \\
        \bottomrule
    \end{tabular}
\end{table*}

\begin{table*}[!t]
    \centering
    \caption{Cross-Benchmark CoT Evaluation}
    \label{tab:cross_benchmark_cot}
    \renewcommand{\arraystretch}{1.3}
    \begin{tabular}{l|c|c|c|c|c|c|c}
        \toprule
        \textbf{Model} & \textbf{CoGenT\_ValB} & \textbf{Describe CoT} & \textbf{PoT} & \textbf{VSR} & \textbf{VSR (SG CoT)} & \textbf{VC Bench} & \textbf{VC Bench (SG CoT)} \\
        \midrule
        Llama-4-Scout-17B-16E-Instruct & 68.47\% & 70.66\% & 40.68\% & 68.82\% & 78.14\% & 53.76\% & 68.08\% \\
        Qwen2.5-VL-7B-Instruct & 61.30\% & 72.70\% & --- & 66.81\% & 70.28\% & 60.54\% & 75.66\% \\
        \bottomrule
    \end{tabular}
    \begin{flushleft}
        \footnotesize \textbf{Note:} VSR refers to the VSR val split. PoT: Program-of-Thought. SG CoT: Scene Graph Chain-of-Thought. Describe CoT is the prompting strategy described in Table VI and methodology sections.
    \end{flushleft}
\end{table*}

The results presented in Table~\ref{tab:ood_comparison} provide clear evidence of the \textbf{generalization gap between supervised fine-tuning (SFT) and GRPO-based reinforcement learning}. On the Depth task, we observe that SFT not only fails to improve performance under out-of-distribution (OOD) conditions, but actually reduces accuracy compared to the base model (56.33\% vs. 59.00\%). This suggests that SFT encourages the model to memorize surface-level linguistic patterns rather than fostering transferable spatial understanding. When test queries shift semantically from "closer to" to "farther from," the SFT model struggles to adapt, highlighting its limited compositional generalization.

In contrast, \textbf{GRPO leads to consistent gains in OOD performance across both tasks}. On the Depth\_OOD setting, GRPO achieves a substantial improvement of \textbf{+19.5\%} over the base model, demonstrating its ability to align vision-language representations in a way that is robust to linguistic variation.

On the Distance task, while both SFT and GRPO improve over the base model in the OOD setting, the degree of degradation from in-distribution performance differs significantly: SFT experiences a \textbf{12.03\%} drop, whereas GRPO sees only a \textbf{3.17\%} reduction. \textbf{This discrepancy further supports the view that GRPO yields stronger generalization, whereas SFT remains sensitive to distributional shifts and prone to overfitting on training-time phrasing.}

\subsection{Chain-of-Thought}

Scene Graph CoT yields consistent gains of \textbf{5\%–15\%} across spatial reasoning tasks relative to direct answering:

\begin{itemize}[noitemsep]
    \item Even \textbf{counting tasks} benefit, despite not requiring explicit relationships.
    \item \textbf{Two-step prompting} outperforms one-step approaches that combine scene graph and answer generation, likely due to \textit{reward hacking}—models shortcutting reasoning for fast answer generation.
\end{itemize}

\begin{table}[h]
    \centering
    \caption{SAT Accuracy}
    \label{tab:vsr_comparison}
    \renewcommand{\arraystretch}{1.3}
    \begin{tabular}{l|c}
        \toprule
        \textbf{CoT Strategy} & \textbf{SAT Val Accuracy} \\
        \midrule
        Llama-4-Scout-17B-16E-Instruct (no CoT) & 69.42\% \\
        + Optical Flow CoT & 70.41\% \\
        + Scene Graph CoT & 74.40\% \\
        + 5-shot Conversational Scene Graph \\ with Optical Flow CoT & \textbf{77.66\%} \\
        \bottomrule
    \end{tabular}
    \begin{flushleft}
        \textbf{Note:} Optical Flow CoT was only utilised when there were 2 images associated with a question in the SAT Dataset.
    \end{flushleft}. 
\end{table}

\begin{table}[!t]
    \centering
    \caption{CoGenT\_ValB Accuracy with and without CoT}
    \label{tab:cogent_temp}
    \renewcommand{\arraystretch}{1.3}
    \setlength{\tabcolsep}{3.0pt} 
    \begin{tabular}{l|c|c}
        \toprule
        \textbf{Model} & \textbf{CoGenT\_ValB} & \textbf{CoGenT\_ValB CoT} \\
        \midrule
        Llama-4-Scout-17B-16E-Instruct & 68.47\% & 70.66\% \\
        Qwen2.5-VL-7B-Instruct & 61.30\% & 72.70\% \\
        \bottomrule
    \end{tabular}
    \begin{flushleft}
        \textbf{Note:} The CoT prompt here followed a 2 prompt strategy where the model was prompted to: ``\textit{Describe the objects in this image: their color, shape, size, location, relative location, reflectance and material (glossy is metallic, matte is rubber).}'' in the first phase, and then in the second phase, the answer to the question was extracted.
    \end{flushleft}
\end{table}

The Optical Flow CoT approach is effective for dynamic reasoning in SAT, enabling accurate predictions on object and egocentric movements by comparing left/right displacements across image pairs. Coupled with Scene Graph CoT and 5-shot conversational Scene Graph CoT prompting improves accuracy further.

\subsection{Ablation Studies}

We evaluated multiple variants and found several methods degrade performance:

\begin{itemize}
    \item \textbf{PoT + Scene Graph}: Combining programmatic reasoning with scene graphs reduced accuracy due to faulty code generation or logic misalignment.
    \item \textbf{Confidence Scores in Scene Graphs}: Including confidence estimates for each object/attribute/relationship added noise and reduced clarity.
    \item \textbf{Depth Maps}: Adding depth maps as a secondary input slightly decreased accuracy; VLMs may not robustly fuse this modality. This modality did not lead to improved accuracy over the baseline Scene Graph CoT for anything but the depth tasks. While it performed better on Distance and Depth tasks, it lagged behind in Relation and Counting tasks creating an overall lower score.
    \item \textbf{CoS and VoT}: These symbolic reasoning methods were inapplicable for tasks like counting or depth estimation, and generally reduced performance.
    \item \textbf{Visual Language Bi-Directional Alignment (VLBA)}: Inspired by SpatialCoT \cite{liu2025spatialcotadvancingspatialreasoning}, this approach asked the model to map coordinates to objects and vice versa, but underperformed compared to Scene Graph CoT.
\end{itemize}

It is also interesting to note that

\subsection{Few-Shot Prompting Results}

Initial few-shot prompting using flat examples (e.g., tuples of \( I, Q, R, SG, A \)) caused the model to repeat prior answers. Improved strategies:

\begin{itemize}
    \item \textbf{Split Answer-Free and Answer-Included Prompts}: Prevents premature answering. 
    \item \textbf{Conversational Format}: Alternating user/assistant messages increases flexibility and grounding, while still aligning the model towards generating scene graphs as part of the reasoning process. Increases accuracy further.
\end{itemize}

\section{Conclusion}

This study provides a comprehensive evaluation of visual language models (VLMs) across spatial reasoning tasks using diverse datasets and benchmarks. We first examined the effectiveness of various Chain-of-Thought (CoT) prompting strategies, finding that simple CoT often fails to enhance performance and may even be detrimental in spatial tasks. In contrast, structured prompting based on scene graphs (SceneGraph CoT) significantly improves spatial interpretation and counting accuracy, revealing the importance of compositional reasoning cues and highlighting the potential of structured prompting in multimodal tasks. Furthermore, utilizing in-context learning via few-shot prompting can further enhance spatial understanding and Question-Answering in VLMs. To this end, we develop a novel Conversational Style Scene Graph CoT prompting. Building on these insights, we applied Group Relative Policy Optimization (GRPO) to fine-tune VLMs on the SAT dataset. GRPO yields robust gains, particularly under out-of-distribution (OOD) conditions, by reinforcing correct reasoning paths rather than memorizing fixed linguistic templates. Compared to supervised fine-tuning (SFT), GRPO offers stronger generalization and alignment between visual and language modalities. While our results show clear benefits within a controlled evaluation setting, future work is needed to assess the robustness of these methods across architectures, modalities, and spatial reasoning variants.

\subsection{Future Work}
Building on these findings, future work will focus on extending CoT frameworks by incorporating attention-guided CoT with scene graphs\cite{zhang2025embodiedvsrdynamicscenegraphguided}, exploring interleaved CoT structures, and refining object detection capabilities to better capture spatial and temporal dynamics. Additionally, we aim to integrate Coconut (Continuous CoT) to maintain reasoning consistency across varying input modalities\cite{hao2024traininglargelanguagemodels}. Fine-tuning on specialized reasoning datasets and leveraging CoT backtracking and reflection will further enhance model robustness. Ultimately, we plan to consolidate these techniques under a unified framework with improved accuracy and interpretability.

While our study focuses on applying GRPO to enhance spatial reasoning in image-based vision-language models, an important direction for future work is extending this approach to video understanding tasks. Recent efforts such as Spatial-R1\cite{ouyang2025spatialr1enhancingmllmsvideo} and VideoChat-R1\cite{li2025videochatr1enhancingspatiotemporalperception} demonstrated that GRPO can be effectively adapted for temporal reasoning, enabling models to reason over object dynamics, motion-based spatial relations, and multi-frame context. This has important implications for embodied decision-making, where tasks like planning, memory, and long-horizon reasoning rely heavily on consistent spatial and temporal understanding.

And recent research suggests that enhancing spatial reasoning may not rely solely on architectural changes, but also on enriching the model's input representations. For example, incorporating depth maps\cite{cai2025spatialbotprecisespatialunderstanding}, 3D positional priors\cite{ma2024spatialpinenhancingspatialreasoning}, or scene reconstructions\cite{chen2024spatialvlmendowingvisionlanguagemodels} can offer explicit spatial cues that are difficult to infer from raw RGB inputs alone. Integrating such structured spatial information into the training pipeline—potentially in conjunction with reinforcement learning—could further improve alignment and generalization in complex multimodal environments. Future work may explore integrating such spatial priors into the training pipeline, enabling models to reason not only over visual content, but also over structured 3D spatial information.

\bibliographystyle{IEEEtran}
\bibliography{references}
\end{document}